%% file: Tutorial_PhysicsTransformers.tex
\documentclass[journal]{IEEEtran}
\usepackage{bm}
\usepackage{booktabs}
\usepackage{float}
\usepackage{amsmath,amsfonts}
\usepackage[noend]{algpseudocode}
\usepackage{algorithm}
\usepackage{array}
\usepackage{multirow}
\usepackage{booktabs}
\usepackage{pifont}
\usepackage[flushleft]{threeparttable}
\usepackage{dsfont}
\usepackage{textcomp}
\usepackage{stfloats}
\usepackage{url}
\usepackage{verbatim}
\usepackage{graphicx}
\usepackage{cite}
\usepackage{graphicx}
\usepackage{lipsum}
\usepackage{placeins}
\usepackage[table]{xcolor}
\usepackage{rotating}
\usepackage{lipsum}
\usepackage{eqparbox}
\usepackage{etoolbox}  
\usepackage[acronym]{glossaries}
\usepackage{colortbl} 
\usepackage{color, colortbl}
\usepackage[inkscapelatex=false]{svg}

\usepackage[hidelinks,colorlinks=true,linkcolor=blue,urlcolor=blue,citecolor=blue]{hyperref}

\usepackage{setspace} 

\usepackage{makecell}

\usepackage{collcell}
\usepackage{pgf}
\usepackage{pgfplots}
\pgfplotsset{compat=1.18}
\usepackage{pgfplotstable}
\usepackage{hhline}
\usepackage{multirow}

\usepackage{tikz}
\usepackage[margin=1in]{geometry}
\usetikzlibrary{positioning, shapes.multipart, arrows.meta}

\def\colorModel{hsb} 

\newcommand\ColCell[1]{
  \pgfmathparse{#1<50?1:0}  
    \ifnum\pgfmathresult=0\relax\color{white}\fi
  \pgfmathsetmacro\compA{0}      
  \pgfmathsetmacro\compB{#1/100} 
  \pgfmathsetmacro\compC{1}      
  \edef\x{\noexpand\centering\noexpand\cellcolor[\colorModel]{\compA,\compB,\compC}}\x #1
  } 
\newcolumntype{E}{>{\collectcell\ColCell}m{0.4cm}<{\endcollectcell}}  

\algnewcommand\algorithmicforeach{\textbf{for each}}
\algdef{S}[FOR]{ForEach}[1]{\algorithmicforeach\ #1\ \algorithmicdo}

\definecolor{LightCyan}{rgb}{0.88,1,1}
\usepackage{todonotes}

\usepackage[hidelinks]{hyperref}
\usepackage[capitalise]{cleveref} 
\begin{document}

\title{Physics-Informed Machine Learning for Transformer Condition Monitoring -- Part I: Basic Concepts, Neural Networks, and Variants}

\author{Jose I. Aizpurua,~\IEEEmembership{Senior Member,~IEEE}

\thanks{J. Aizpurua is with the University of the Basque Country (UPV/EHU), Faculty of Informatics, Department of Computer Science and Artificial Intelligence, Donostia - San Sebastian, Spain.}
\thanks{978-84-XX-XXXXX-X /25 © 2025 Red iNtransf}
\thanks{\textcolor{red}{“© 2025 IEEE.  Personal use of this material is permitted.  Permission from IEEE must be obtained for all other uses, in any current or future media, including reprinting/republishing this material for advertising or promotional purposes, creating new collective works, for resale or redistribution to servers or lists, or reuse of any copyrighted component of this work in other works.”}}
}

\markboth{2025 8th International Advanced Research Workshop on Transformers (ARWtr)  –   Baiona - Spain, (12)13-15 October 2025}%
{Shell \MakeLowercase{\textit{et al.}}: Bare Demo of IEEEtran.cls for IEEE Journals}

\maketitle

\begin{abstract} 
Power transformers are critical assets in power networks, whose reliability directly impacts grid resilience and stability. Traditional condition monitoring approaches, often rule-based or purely physics-based, struggle with uncertainty, limited data availability, and the complexity of modern operating conditions. Recent advances in machine learning (ML) provide powerful tools to complement and extend these methods, enabling more accurate diagnostics, prognostics, and control. In this two-part series, we examine the role of Neural Networks (NNs) and their extensions in transformer condition monitoring and health management tasks. This first paper introduces the basic concepts of NNs, explores Convolutional Neural Networks (CNNs) for condition monitoring using diverse data modalities, and discusses the integration of NN concepts within the Reinforcement Learning (RL) paradigm for decision-making and control. Finally, perspectives on emerging research directions are also provided.
\end{abstract}


\section{Introduction}
\label{sec:Intro}

\IEEEPARstart{T}{ransformers} are key components in the reliable operation of the power grid. Consequently, monitoring their health is essential to avoid costly failures and unplanned outages, especially in critical facilities such as nuclear power plants. Over the past decade, I have developed and deployed machine learning (ML) solutions for transformer diagnostics and prognostics in industrial contexts \cite{Aizpurua_22}.

Most of the developed solutions have been focused on Prognostics \& Health Management (PHM) applications. PHM is a health management paradigm that integrates different predictive methodologies to improve the reliability of engineering components and systems \cite{Aizpurua_SI_TrafoMagazine_22}. PHM is at the heart of modern condition monitoring technology, where sensor data, physics knowledge, and engineering expertise are combined to develop anomaly detection, diagnostics, prognostics, and maintenance planning tools.

Under the umbrella of PHM methods for electrical transformers, several data-driven and hybrid approaches have been developed, including a data analytics suite \cite{Analytics_2017}, insulation lifetime prognostics through ML and state-space filtering \cite{Aizpurua_TII}, and diagnostic models using Dissolved Gas Analysis (DGA) data with Bayesian Networks \cite{Aizpurua_2018}. The latter was further enhanced through the integration of uncertainty-aware ensemble strategies \cite{Aizpurua_21}. Engineers often need to simplify the decision-making process through a single health indicator, and in this direction, a composite health index was also proposed to integrate predictions from multiple transformer subsystems \cite{Aizpurua_2019}.

The combination of ML and physics-based models, together with uncertainty quantification (UQ), has been a central element in most of these solutions. UQ is essential for decision-making, particularly in prognostics, where Remaining Useful Life (RUL) predictions must incorporate uncertainty estimates. Using probabilistic forecasting methodologies, a hybrid physics–ML method for transformer ageing prediction under renewable energy operation was developed in \cite{Aizpurua23}. The physics-based model served as the baseline, while the probabilistic component learned and corrected its residual bias. A comprehensive study of these probabilistic forecasting methods and their validation can be found in \cite{Aizpurua_22}.

Beyond insulation ageing, transformer magnetization and inrush current phenomena have also been investigated. Based on the Jiles–Atherton model for hysteresis curve modelling, a metaheuristics-based approach was proposed to estimate the underlying partial differential equation (PDE) parameters \cite{Ugarte_24}. This framework was later extended toward transformer inrush minimization strategies, integrating the hysteresis model with a transformer electrical model in a Reinforcement Learning (RL) setup for optimal control \cite{RL_Jone}.

Motivated by the need for cost-effective and non-invasive monitoring, recent work has also explored the use of acoustic signals and ML for On-Load Tap Changer (OLTC) monitoring \cite{Secic25}. This line of research demonstrates the potential of combining signal processing and data-driven techniques for transformer subsystem diagnostics.

Given the wide range of ML-based solutions proposed for transformer health management, this tutorial aims to provide a unified and accessible overview of Neural Network (NN)-based methods. NNs are chosen as the central modelling framework because many of their variants, such as Convolutional NNs and Reinforcement Learning agents, are natural extensions of the same underlying principles.

The increasing penetration of Renewable Energy Sources (RESs) has also affected transformer health. Their stochastic nature and the variability they introduce to grid operations have motivated the development of complementary solutions, including: (i) feature selection for improved forecasting \cite{Ramirez_24}; (ii) surrogate modelling for transformer thermal dynamics \cite{Aizpurua_23}; and (iii) analysis of the impact of extreme events on transformer ageing \cite{SanchezPozo_25}. However, these applications and broader maintenance planning approaches (e.g., \cite{Aizpurua_17}) are beyond the scope of this tutorial.

This paper introduces the fundamentals of ML methods for transformer condition monitoring, with a focus on Neural Networks and their extensions. Part~I  presents the foundations of Neural Networks, including fully connected (feedforward) architectures, Convolutional Neural Networks (CNNs), and their integration within the Reinforcement Learning (RL) paradigm for transformer control. Part~II builds upon these foundations by introducing Physics-Informed Neural Networks (PINNs) and Bayesian methods for uncertainty quantification. Together, the two parts aim to provide researchers and practitioners with a coherent framework for developing data-efficient, physics-consistent, and uncertainty-aware models for transformer health monitoring and asset management.

The remainder of this document is organized as follows. Section~\ref{Sec:NeuralNetworks} introduces the basic concepts of Neural Networks. Section \ref{Sec:CNN} presents Convolutional Neural Networks for structured data. Section~\ref{Sec:RL_Trafo} discusses Reinforcement Learning for transformer control, and finally, Section~\ref{s:Conclusions} concludes with perspectives for future research.

\section{Neural Networks}
\label{Sec:NeuralNetworks}

\subsection{Basic Concepts}

Artificial Neural Networks (NNs) are computational models inspired by the structure and functioning of biological neurons \cite{Scardapane2024}. The most basic architecture, known as a feedforward neural network or multilayer perceptron (MLP), consists of an input layer, one or more hidden layers, and an output layer, as illustrated in Fig.~\ref{fig:NN_basic}.

\input{NN_Fig.tex}

Each neuron receives an input vector $X$, applies a linear transformation with weights $W$ and bias $b$, and passes the result through a nonlinear activation function $\sigma(\cdot)$. Formally, a fully connected (FC) layer is defined as:

\begin{equation}
    FC(X)=\sigma(XW+b)
\end{equation}

\noindent where $X \in \mathbb{R}^{n \times c}$ is the input with $n$ samples of dimension $c$, $W \in \mathbb{R}^{c \times c'}$ is the weight matrix, and $b \in \mathbb{R}^{c'}$ is the bias vector. The parameter $c'$ determines the number of neurons (or the width) in the layer. The nonlinear activation function $\sigma(\cdot)$ allows the network to approximate complex, nonlinear relationships.

Among the various activation functions used in practice (e.g., ReLU, sigmoid, tanh), this tutorial employs the softplus function,

\begin{equation}
    \sigma(s)=log(1+exp(s))
\end{equation}

\noindent which is smooth, always positive, and avoids dead (or inactive) neurons that can occur with other activation functions such as ReLU.

Training a Neural Network involves optimizing the parameters $\{W_k, b_k\}_{k=1}^{L}$ across $L$ layers to minimize a loss function that quantifies the difference between predictions and true outputs. For regression tasks, the mean squared error (MSE) is a common choice:

\begin{equation}
\label{eq:Loss}
\mathcal{L} = \frac{1}{n}\sum_{i=1}^{n} \big(y_i - f(x_i)\big)^2,
\end{equation}

\noindent where $f(x_i)$ is the network prediction for the input $x_i$. The parameters are optimized through gradient-based algorithms such as stochastic gradient descent (SGD), with gradients efficiently computed using the backpropagation algorithm.

Backpropagation applies the chain rule of calculus to propagate the loss gradient backward through the layers, allowing all network parameters to be updated simultaneously. This mechanism enables Neural Networks to efficiently learn complex input–output mappings from data.

\subsection{Learning Tasks. Classification and Regression}

Neural Networks can be tailored to different supervised learning tasks, primarily classification and regression.

\begin{itemize}
\item \textbf{Classification:} The goal is to assign an input to one of several categories. In transformer condition monitoring, an example is fault type identification from Dissolved Gas Analysis (DGA). Traditional methods such as Duval’s triangle rely on threshold rules, whereas NNs can learn probabilistic and nonlinear decision boundaries, improving fault classification accuracy.
\item \textbf{Regression:} The network predicts continuous values. For transformers, regression can be used to predict oil temperature or estimate remaining useful life (RUL) of insulation based on operational history.
\end{itemize}

The choice of the output activation function and loss function depends on the problem type. For instance, classification problems commonly employ a softmax output with categorical cross-entropy loss:

\begin{equation}
\label{eq:Loss_CrossEntropy}
\mathcal{L} = -\sum_{k=1}^{K} y_k \log(p_k),
\end{equation}

\noindent where $p_k$ is the predicted probability of class $k$, and $y_k$ is the one-hot encoded true label. In contrast, regression tasks typically use a linear output layer and the MSE loss defined in Eq.~(\ref{eq:Loss}).

\subsection{Transformer Applications}

Neural Networks have been successfully applied to a range of transformer asset management and condition monitoring tasks within the PHM framework. Some representative applications include:

\begin{itemize}
    \item Fault diagnostics from sensor data (e.g., DGA) using fully connected NNs to identify gas ratios and fault categories \cite{Aizpurua_2018}.
    \item Prognostics and Remaining Useful Life (RUL) prediction of insulation degradation from operational data \cite{Aizpurua_TII}.
    \item Ensemble learning strategies that combine multiple models trained on different datasets to improve robustness and accuracy in diagnostics \cite{Aizpurua_21}.
\end{itemize}

Neural Networks provide a flexible and expressive modeling framework capable of capturing complex nonlinear relationships. However, they also exhibit limitations, including a tendency to overfit when training data are scarce, limited interpretability, and the need for careful regularization to ensure generalization \cite{Scardapane2024}.

Motivated by the need to better handle structured and high-dimensional data such as images and time–frequency representations, specialized architectures have been proposed, such as Convolutional Neural Networks (CNNs), which are introduced in the following section.

\section{Convolutional Neural Networks}
\label{Sec:CNN}

\subsection{Motivation: Why Specialized Architectures?}
\label{ss:IntroCNN}

In transformer monitoring, data come in diverse forms: gas concentration vectors, temperature time series, high-frequency acoustic signals, or thermal images. Conventional feedforward Neural Networks (MLPs) process each input feature independently and thus fail to exploit local structure or spatiotemporal correlations. However, in many cases, these correlations carry essential diagnostic information, such as bursts in the frequency domain, thermal hotspots, or waveform patterns, crucial for identifying specific fault conditions.

Table~\ref{Table:DataTypes} summarizes typical transformer monitoring data sources, their structure, and representative patterns.

\begin{table}[!htb]
\scriptsize
\label{Table:DataTypes}
\begin{center}
\caption{Data types used in transformer monitoring and their characteristic patterns}
\setlength{\tabcolsep}{3.000pt}
\begin{tabular}{|l|l|l|}
\hline
\textbf{Source} & \textbf{Format} & \textbf{Patterns} \\
\hline
DGA & Low-dimensional vectors & Gas ratios \\
Load, temperature & 1D time series & Trends, seasonal effects \\
OLTC acoustics, vibration & Waveforms (20\,kHz–1\,MHz) & Time-frequency bursts \\
IR/thermal imaging & 2D images & Spatial hot-spots \\
\hline
\end{tabular}
\end{center}
\end{table}
\normalsize

Convolutional Neural Networks (CNNs) are designed to capture such localized and hierarchical patterns. Instead of fully connected layers, CNNs employ learnable filters (kernels) that slide across the input domain, detecting meaningful features, such as edges, bursts, or hotspots, independent of their position. These spatially shared filters enable efficient learning from high-dimensional, structured data \cite{Zewen_22}.

\subsection{CNN Fundamentals: Local Filters and Feature Hierarchies}
\label{ss:CNN_Fundamentals}

Consider an input tensor $X \in \mathbb{R}^{h \times w \times c}$, representing an image of height $h$, width $w$, and $c$ channels (e.g., grayscale or RGB). A convolutional filter $K \in \mathbb{R}^{k_h \times k_w \times c}$ produces an output feature map $Y$ via the convolution operation:

\begin{equation}
Y_{i,j}^{(k)} = \sigma\left( \sum_{u,v,c} K_{u,v,c}^{(k)} \cdot X_{i+u,j+v,c} + b^{(k)} \right),
\end{equation}

\noindent where $\sigma(\cdot)$ is a nonlinear activation function (typically ReLU), and $b^{(k)}$ is a bias term for the $k$-th feature map.

As an illustrative example, a $3 \times 3$ filter can be defined as:

\begin{equation}
K = \begin{bmatrix}
-1 & -1 & -1 \\
 0 &  0 &  0 \\
 1 &  1 &  1 \\
\end{bmatrix}
\end{equation}

This kernel acts as a vertical edge detector, emphasizing horizontal transitions in the input image. The convolution is performed by sliding the kernel over the input region, computing element-wise products, summing them, and applying the activation function:

\begin{equation}
Y_{i,j} = \sigma\left( \sum_{u=0}^{2} \sum_{v=0}^{2} K_{u,v} \cdot X_{i+u, j+v} + b \right).
\end{equation}

During training, CNNs automatically learn a diverse set of filters that respond to different features—edges, textures, or frequency transitions. Shallow layers capture low-level details, while deeper layers learn higher-level abstractions such as characteristic OLTC acoustic patterns or thermal fault signatures.

Key benefits of CNNs include:

\begin{itemize}
    \item Parameter efficiency. The same kernel detects a feature across the entire input, reducing parameters and improving generalization.
    \item Spatio-temporal information. Filters operate on spatial or temporal neighborhoods, detecting meaningful patterns.
    \item Hierarchical abstraction. Stacked layers construct features from low-level edges to complex patterns.
    \item Pooling operations. Downsampling (e.g., max-pooling)  reduces dimensionality while preserving dominant features.
\end{itemize}

Although CNNs are commonly associated with 2-D image data, they are also applicable to 1-D signals such as vibration or acoustic recordings using 1-D convolutions. Alternatively, transforming time series into 2-D time–frequency representations (e.g., spectrograms) allows CNNs to leverage their full spatial feature extraction capabilities.

\subsection{From Time Series to Images: Spectrograms for Acoustic Monitoring}

In the context of transformer diagnostics, particularly On-Load Tap Changer (OLTC) monitoring, acoustic signals contain information about electrical and mechanical events. The frequency content of these signals varies over time (i.e., they are non-stationary), which motivates the use of time–frequency analysis.

A widely used method for this purpose is the (discrete) Short-Time Fourier Transform (STFT), which decomposes a signal into overlapping time-localized frequency components:

\begin{equation}
S(t, f) = \left| \sum_{n=0}^{L-1} x[n + t] \cdot w[n] \cdot e^{-j 2\pi f n / L} \right|,
\end{equation}

\noindent where $x[n]$ is the discrete-time input signal obtained with respect to a sampling rate $F_s$, $w[n]$ is a window function (e.g., Hamming or Hann) of length $L$, $t$ denotes the time index (window shift), $f$ is the frequency bin index. The window length $L$ determines the temporal resolution (\textit{i.e.} L/$F_s$ seconds). For simplicity, the hop size (window shift step), is omitted.

By sliding the analysis window over time, the STFT yields a two-dimensional representation, known as the spectrogram:

\begin{equation}
\text{Spectrogram}(t, f) = 20 \log_{10} \left( |S(t, f)| \right).
\end{equation}

Log-scaling enhances perceptual relevance and dynamic range. In practice, Mel-scaled spectrograms are often employed to mimic human auditory perception, particularly effective for identifying low-frequency variations critical to OLTC event detection.

This transformation enables CNNs to process audio data as images, learning discriminative patterns from the spectrograms for event classification or anomaly detection.

\subsection{Case Study: CNN-Based OLTC Acoustic Monitoring}

In \cite{Secic25}, a CNN architecture was developed to detect OLTC operational states from acoustic recordings sampled at 48\,kHz. Fig.~\ref{fig:ElinAudio_CNN} illustrates the CNN structure.

\begin{figure}[!htb]
\centering
\includegraphics[width=\columnwidth]{Images/ELIN_Test_cnn.png}
\caption{Schematics of the CNN architecture. The numbers $a \times b \times c$ inside the Conv2D boxes indicate the kernel size and the number of filters \cite{Secic25}.}
\label{fig:ElinAudio_CNN}
\end{figure} 

\begin{figure*}[!htb]
\centering
\includegraphics[width=1.8\columnwidth]{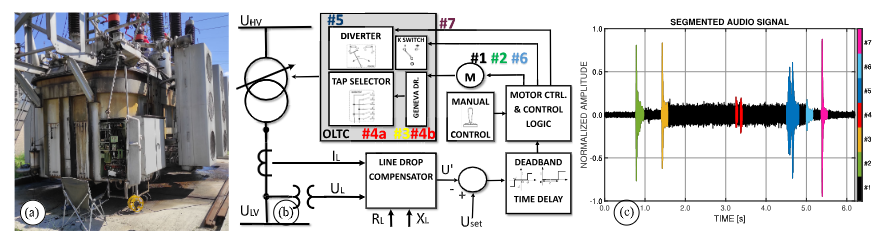}
\caption{(a) ELIN test environment (b) control scheme with acoustic sources; (c) segmented audio signal \cite{Secic25}.}
\label{fig:ElinAudio}
\end{figure*}

The model input consists of Mel spectrograms corresponding to 360\,ms audio segments (17$\times$1024 samples). The CNN classifies each segment into one of the OLTC states defined in Table~\ref{table:OLTC_Trafo_States}. Fig.~\ref{fig:ElinAudio}(a) shows the test environment, Fig.~\ref{fig:ElinAudio}(b) depicts the voltage regulation control scheme, and Fig.~\ref{fig:ElinAudio}(c) presents the segmented acoustic signature corresponding to one OLTC transition.

\begin{table}[!hbtp]
	\centering
    \scriptsize
	\caption{Segmentation of transformer OLTC states.}
	\label{table:OLTC_Trafo_States}
	\begin{tabular}{
   >{\centering\arraybackslash}m{.6cm}|
   >{\centering\arraybackslash}m{6cm}}
		\hline
		{\textbf{State}} & {\textbf{Description}}\\ \hline	
		{\#1} & {Idle state and motor operation}\\ \hline	
  {\#2} & {Motor starts}\\ \hline	
  {\#3} & {Geneva drive activation}\\ \hline	
  {\#4} & {Tap selector stops and end of Geneva drive}\\ \hline	
  {\#5} & {Diverter switch operation}\\ \hline	
  {\#6} & {Motor stops}\\ \hline	
  {\#7} & {Events associated with the mechanical braking process}\\ \hline	
	\end{tabular}	
\end{table}

The proposed approach achieved over 98\% classification accuracy on unseen audio signals, demonstrating the strong feature extraction capability of CNNs for high-dimensional acoustic data. Compared to traditional feature-engineered machine learning models, CNNs exhibited superior adaptability and generalization across varying acoustic environments \cite{Secic25}.

The model’s robustness was further evaluated under both synthetic and real noise conditions. While the CNN maintained high accuracy in the presence of synthetic noise, real-world tests revealed sensitivity to background humming, requiring a pre-processing denoising step for optimal performance. These findings underscore the importance of noise-robust preprocessing in practical deployments.

CNNs thus extend Neural Networks to structured data, enabling hierarchical feature extraction from complex sensor signals. Their ability to generalize across noise conditions and learn diagnostic representations makes them a key enabler for future transformer health monitoring solutions.

\section{Reinforcement Learning for Transformer Control}
\label{Sec:RL_Trafo}

Transiting towards Prognostics and Health Management (PHM) solutions that integrate predictive information requires coupling condition assessment with optimal decision-making. One example would be the use of insulation ageing predictions to plan cost-effective maintenance actions. In this direction, Reinforcement Learning (RL) provides a framework that enables adaptive, data-driven control by learning from interaction with the system \cite{Sutton2015}. Rather than focusing on maintenance scheduling, this section illustrates how RL can be used for transformer actuation control, specifically for minimizing inrush current during energization—a phenomenon that directly contributes to accelerated ageing.

\subsection{Motivation: Adaptive Control with Neural Policies}

Traditional control strategies for transformers, such as fixed switching sequences, often rely on static settings. While these methods are effective under nominal steady-state conditions, they cannot easily adapt to varying load profiles, fluctuating grid conditions, or complex operational trade-offs. Moreover, these rule-based systems rarely integrate real-time health signals or system feedback in a principled manner.

RL provides a data-driven alternative for sequential decision-making. Instead of pre-defined rules, an RL agent learns optimal control strategies through interaction with an environment, which may be a detailed simulator or, eventually, a real operating environment. RL agents can dynamically adapt to context, achieving objectives such as minimizing inrush currents, reducing thermal stress, or improving lifetime utilization. Namely they learn to take actions that maximize a cumulative reward function, capturing both short-term performance and long-term effects. Neural networks are typically used to represent the agent’s decision policy, making RL another extension of NNs toward control.

\begin{figure}[ht]
    \centering
    \includegraphics[width=1\linewidth]{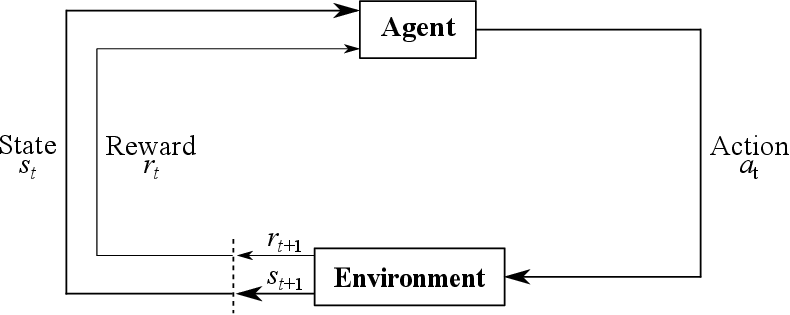}
        \caption{Agent–environment interaction in reinforcement learning. The agent learns an optimal policy through repeated interaction with the environment.}
    \label{fig:RL_basecs}
\end{figure}

Fig.~\ref{fig:RL_basecs} illustrates the classical RL agent–environment interaction loop. At each time step $t$, the agent observes the environment state $s_t$, selects an action $a_t$, and receives a scalar reward $r_{t+1}$ and a new state $s_{t+1}$ from the environment. The goal is to learn a policy $\pi(a_t|s_t)$ that maximizes the expected cumulative reward.

The motivations to use RL in transformer health and control include: (i) the ability to learn from interaction without requiring labeled datasets or expert heuristics, (ii) the use of simulators (environment) to explore long-term effects safely, and (iii) the potential to balance multiple, sometimes conflicting, objectives via the design of custom reward functions (e.g., mechanical wear versus voltage quality).

\subsection{RL Fundamentals: Control as a Markov Decision Process}

Formally, an RL problem is modeled as a \textit{Markov Decision Process} (MDP), defined by the tuple $(\mathcal{S}, \mathcal{A}, P, R, \gamma)$, where:

\begin{itemize}
    \item \( \mathcal{S} \): Set of possible system states (e.g., transformer fluxes, grid voltage, ambient temperature).
    \item \( \mathcal{A} \): Set of available actions (e.g., energization angle, tap changer positions).
    \item \( P(s'|s,a) \): State transition probabilities, defining system dynamics.
    \item \( R(s,a) \): Reward function that quantifies the desirability of each action.
    \item \( \gamma \in [0,1] \): Discount factor that determines the importance of future rewards.
\end{itemize}

The agent selects actions according to a policy \( \pi(a_t|s_t) \), often represented by a Neural Network, to maximize the expected cumulative reward over time. In most deep RL implementations, Neural Networks serve as policy approximators (actor), value estimators (critic), or both. These function approximators allow generalization across high-dimensional state spaces.

\subsection{Case Study: Minimizing Inrush Current via Neural Control}

A critical application of RL in transformers is the mitigation of inrush current during energization. Due to residual magnetic flux in the core, inappropriate circuit breaker closing instants can cause inrush currents several times above nominal levels, leading to mechanical stress, false protection operation, and voltage dips. In this context, classical rule-based control is often insufficient, and this makes it a natural candidate for RL methods.

To address this, an RL agent can be trained in simulation to determine the optimal energization instant (i.e., circuit breaker closing angle) that minimizes inrush current. The environment includes a transformer model receiving as inputs the three-phase remanent flux values \((\phi_1, \phi_2, \phi_3)\) and the selected switching angle (action). It outputs the peak inrush current \( I_{\text{max}} \), which is used to compute the reward:

\begin{equation}\label{eq:reward}
r =
\begin{cases} 
    -I_{\text{max}}, & \text{if } I_{\text{max}} > 1, \\
    1 - I_{\text{max}}, & \text{otherwise.}
\end{cases}
\end{equation}

\begin{figure}[!htb]
    \centering
    \includegraphics[width=0.8\linewidth]{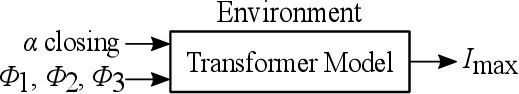}
    \caption{Transformer simulation environment. Inputs: remanent fluxes and closing angle. Output: peak inrush current \cite{RL_Jone}.}
    \label{fig:Environment}
\end{figure}

This incentivizes the agent to find energization strategies that keep inrush below the rated current (1 pu). The diagram in Fig.~\ref{fig:Environment} shows the inputs and output of the environment.

\begin{figure}[ht]
    \centering
    \includegraphics[width=1\linewidth]{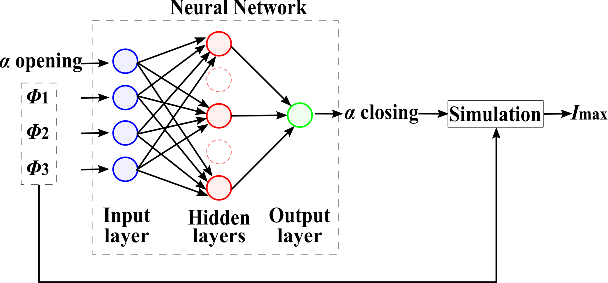}
    \caption{Evaluation stage: trained policy maps state to closing angle. Simulation outputs inrush current \cite{RL_Jone}.}
    \label{fig:eval_exp}
\end{figure}

After training, the agent receives state observations (e.g., flux and timing) and outputs the optimal closing angle that minimizes inrush. The learned policy is evaluated in the simulation environment, as shown in Fig.~\ref{fig:eval_exp}.

\begin{figure}[!htb]
    \centering
    \includegraphics[width=1\linewidth]{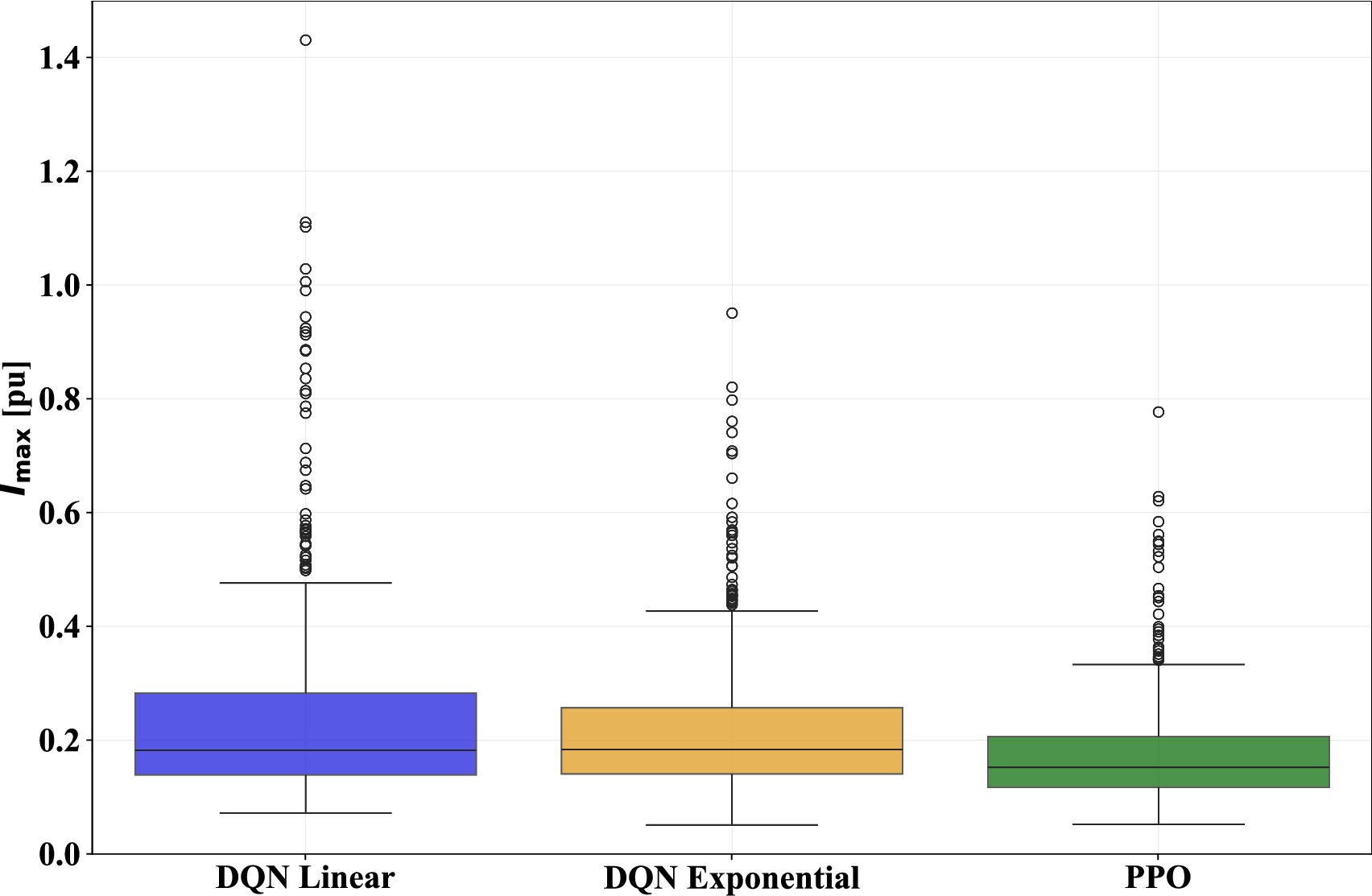}
    \caption{Distribution of peak inrush currents for different RL algorithms. PPO yields the lowest mean and variance, achieving safer energization behavior \cite{RL_Jone}.}
    \label{fig:boxplot}
\end{figure}

In \cite{RL_Jone}, multiple RL algorithms were benchmarked, including Deep Q-Networks (DQN) with linear and exponential $\epsilon$-greedy decay, and Proximal Policy Optimization (PPO). Fig.~\ref{fig:boxplot} compares the distribution of peak inrush currents for these approaches. While DQN with linear decay occasionally exceeds the rated limit, PPO and DQN with exponential decay show more consistent and safer performance.

Reinforcement Learning extends Neural Networks beyond predictive modeling into sequential decision-making, where the focus is not only on estimating states but also on learning control strategies that optimize long-term objectives. For transformers, this case study demonstrates how neural agents can effectively control transformer behavior to minimize inrush current.

As high-fidelity simulators and hybrid physics–ML environments become more available, RL-based control will increasingly integrate physical models and uncertainty estimates. Such hybrid models can enable safe RL, where operational constraints and physics laws guide exploration, paving the way toward autonomous and health-aware transformer control.

\section{Conclusions}
\label{s:Conclusions}

In this first part of the tutorial, we have provided an overview of machine learning solutions for transformer condition monitoring based on Neural Networks (NNs) and their main variants. The focus was placed on understanding the basic building blocks of NNs, their adaptation to structured data through Convolutional Neural Networks (CNNs), and their integration into control-oriented frameworks via Reinforcement Learning (RL).

NNs serve as the foundation of most deep learning architectures and provide a powerful means of learning nonlinear relationships directly from data. CNNs extend these capabilities by efficiently processing structured signals, such as acoustic and thermal measurements, enabling robust event detection and feature extraction for On-Load Tap Changer (OLTC) monitoring. RL, on the other hand, generalizes the role of NNs from prediction to decision-making, allowing control policies to be learned directly from interaction with the system, as demonstrated in the inrush current minimization case study.

Future research will likely advance in several directions. Audio-based monitoring shows strong potential for cost-effective condition assessment, though improving robustness to noise remains an open challenge. Another promising direction is the integration of self-supervised learning to enable adaptive diagnosis of previously unseen operating patterns. For Reinforcement Learning, combining decision-making strategies with Remaining Useful Life (RUL) prediction models can pave the way for predictive, cost-effective maintenance scheduling.

Part~II of this series will extend these Neural Network approaches toward physics-informed methods and uncertainty quantification, essential components for robust decision-making under sparse or uncertain data.

Together, Parts I and II aim to provide a coherent view of how emerging AI paradigms, ranging from deep neural networks to physics-informed and probabilistic models, can contribute to the next generation of reliable, explainable, and data-efficient solutions for power transformer monitoring and asset management.

\section*{Acknowledgements}

I am grateful to all my collaborators and mentors, especially Vic Catterson, Stephen McArthur, and Brian Stewart. I would also like to thank Xose M. Lopez-Fernandez for the invitation to present this tutorial at the ARWTr 2025 conference. In this Part I, I would like to acknowledge my collaborators on OLTC audio monitoring: Adnan Secic, Eñaut Muxika, Unai Garro, and Igor Kuzle, as well as on RL based solutions, including Manex Barrenetxea and Jone Ugarte.

This work has been partially funded by the Spanish State Research Agency through the Ramón y Cajal Fellowship (grant No. RYC2022-037300-I) and Proyectos Generación de Conocimiento (grant No. PID2024-156284OA-I00), co-funded by MCIU/AEI/10.13039/501100011033 and FSE+.

\bibliographystyle{IEEEtran}

\bibliography{IEEEabrv,myBib}

\end{document}

%% file: NN_Fig.tex
\begin{figure}[ht]
\centering
\begin{tikzpicture}[scale=0.7,shorten >=1pt,->,draw=black!60, node distance=2.2cm]

  \tikzstyle{input neuron}  =[circle, draw=black!80,
                              fill=blue!20,  minimum size=20pt, inner sep=0pt]
  \tikzstyle{hidden neuron} =[circle, draw=black!80,
                              fill=orange!20, minimum size=20pt, inner sep=0pt]
  \tikzstyle{output neuron} =[circle, draw=black!80,
                              fill=green!20, minimum size=20pt, inner sep=0pt]
  \tikzstyle{loss block}    =[rectangle, draw=red!80,
                              fill=red!20, minimum width=1.6cm,
                              minimum height=1.0cm, align=center]
  \tikzstyle{annot}         =[text width=5em, text centered]

  \foreach \name/\y in {1/1, 2/2, 3/3}
    \node[input neuron] (I-\name) at (0,-\y) {};

  \foreach \name/\y in {1/0.5, 2/1.5, 3/2.5, 4/3.5}
    \node[hidden neuron] (H-\name) at (2,-\y) {$\sigma$};

  \foreach \name/\y in {1/1, 2/2}
    \node[output neuron] (O-\name) at (4,-\y-0.5) {};

  \node[loss block, right=1.8cm of O-1.center] (Loss) {Loss\\ (MSE / CE)};

  \foreach \i in {1,...,3}
    \foreach \h in {1,...,4}
      \draw (I-\i) -- (H-\h);

  \foreach \h in {1,...,4}
    \foreach \o in {1,2}
      \draw (H-\h) -- (O-\o);

  \foreach \o in {1,2}
    \draw (O-\o) -- (Loss);

  \node[annot, above of=I-1, node distance=1.1cm] {Input\\Layer};
  \node[annot, above of=H-1, node distance=1.1cm] {Hidden\\Layer};
  \node[annot, above of=O-1, node distance=1.1cm] {Output\\Layer};

\end{tikzpicture}
\caption{Multilayer Perceptron (MLP) with activation functions in the hidden layer and a loss function at the output.}
\label{fig:NN_basic}
\end{figure}